\documentclass[sigconf,nonacm]{acmart}

\copyrightyear{2023} 
\acmYear{2023} 
\setcopyright{acmlicensed}
\acmConference[ICONS '23]{International Conference on Neuromorphic Systems}{August 1--3, 2023}{Santa Fe, NM, USA}
\acmBooktitle{International Conference on Neuromorphic Systems (ICONS '23), August 1--3, 2023, Santa Fe, NM, USA}
\acmPrice{15.00}
\acmDOI{10.1145/3589737.3606009}
\acmISBN{979-8-4007-0175-7/23/08}

\usepackage[disable]{todonotes}
\usepackage{graphicx}
\usepackage{bm}
\usepackage{wrapfig}
\usepackage[separate-uncertainty = true]{siunitx}
\DeclareSIUnit{\million}{~\text{M}}
\DeclareSIUnit{\kilo}{~\text{K}}
\usepackage{caption}
\usepackage[subtle]{savetrees}
\begin{document}

\title{
Beyond Weights: Deep learning in Spiking Neural Networks with pure synaptic-delay training
}


\author{Edoardo W. Grappolini}
\email{edoardo.grappolini@stud.unifi.it}
\affiliation{%
  \institution{Università degli Studi di Firenze}
  \country{Italy}
}

\author{Anand Subramoney}
\email{anand.subramoney@rhul.ac.uk}
\affiliation{%
  \institution{Royal Holloway, University of London}
  \country{UK}
}


\begin{abstract}
    Biological evidence suggests that adaptation of synaptic delays on short to medium timescales plays an important role in learning in the brain.
    Inspired by biology, we explore the feasibility and power of using synaptic delays to solve challenging tasks even when the synaptic weights are not trained but kept at randomly chosen fixed values.
    We show that training ONLY the delays in feed-forward spiking networks using backpropagation can achieve performance comparable to the more conventional weight training.
    Moreover, further constraining the weights to ternary values does not significantly affect the networks' ability to solve the tasks using only the synaptic delays.
    We demonstrate the task performance of delay-only training on MNIST and Fashion-MNIST datasets in preliminary experiments.
This demonstrates a new paradigm for training spiking neural networks and sets the stage for models that can be more efficient than the ones that use weights for computation.
\end{abstract}





\maketitle

\section{Introduction}\label{sec:introduction}

Spiking neural networks (SNNs)~\citep{maass1997networks} are biologically inspired neuron models that have recently become increasingly popular for deep learning use cases due to their potential for extreme energy efficiency on neuromorphic hardware. 
In these models, the complex electrochemical dynamics of a biological synapse are modelled in the connection between neurons and (typically) weights parameters. 
Neuroscientific literature classically focuses on the neuron and synaptic strength as the only factor of learning; this was because of several factors, the most trivial of which is that electrical activity is a relatively easy measurement of cellular and brain activity. 
Recent literature \citep{fields_Brain_2020} suggests, however, that other types of cells could contribute to computations and learning in the brain.
Glia cells, especially Oligodendrocytes, have been shown to be activated through learning processes \citep{monje_2018, huges_2018}. 
At a very high level, they work by wrapping the axon in a myelin sheath which can control the electrical signal speed through synapses and as a consequence, the time of the spike reception.

The role and importance of exact spike times are also considered in many studies \citep{gollisch_2008, thorpe_1990, brette_2015} and the hypothesis is also reinforced by simulative works~\citep{guo_2021}. 


In this work, we explore the potential of computations being a direct consequence of synaptic delays in a network of spiking neurons. 
Although the idea of computations through delays comes from nature, our techniques come from the current technical literature in spiking neural networks, and we train our models through backpropagation. 
More precisely, we use an SNN based on SLAYER~\citep{slayer} and compare training only the synaptic delays with the conventional procedure of training weights of the network. 
We show that training just the synaptic delays can lead to competitive task performance on deep learning benchmarks -- specifically MNIST and Fashion-MNIST.


\section{Related work}\label{sec:related-work}

Many previous works explore the possibility of using time coding schemes and precise spike time learning in spiking neural networks as well as training the delays in an SNN.
Examples include \citep{mostafa_supervised_2018, goltz_Fast_2021, s4nn} that combines time coding and backpropagation for different types of neurons (IF, LIF). 
DL-ReSuMe \citep{DLresume}, as an extension of ReSuMe \citep{resume} introduces the concept of delay training to improve performance and reduce weight adjustments, utilizing a supervised learning rule that is not backpropagation based. 
\citet{hazan2022memory} is perhaps the most closely related work to ours, where only the delays of a weightless spiking neural network are trained using an STDP unsupervised learning rule to create latent representations of the MNIST dataset, which are then classified using a linear classifier. 
SLAYER \citep{slayer} uses pseudo-derivatives to train axonal delays and weights using backpropagation with a spike response model (SRM) of spiking neurons.
Our work was heavily influenced by SLAYER, although we train synaptic rather than axonal delays and explore the case where weights aren't trained.
To our knowledge, we are the first to show that pure delay training using backpropagation with surrogate gradient method can achieve comparable performance to weight training.


\section{Methods}\label{sec:methods}

\subsection{Spike response model (SRM)}
We utilize the spike response model (SRM) and all the parameters were trained using surrogate gradients as in~\citep{slayer}.
This includes the synaptic weights and delays. 
For the convenience of the reader, we describe the notation necessary for the comprehension of our work. 

Let $s_i(t) = \sum_f \delta(t-t_i^{(f)})$ be one of a series of spike trains that reaches a neuron;  $t_i^{(f)}$ is the time of the $f^{th}$ spike of the $i^{th}$ input.
Let $\epsilon(\cdot)_d$ be a spike response kernel that also takes into consideration the \textit{axonal} delay. 
Then the membrane potential of the neuron that we are taking into consideration is: 
\begin{align*}
u(t) &= \sum_i w_i(\epsilon_d\ast s_i)(t) + (\nu\ast s) = \bm{w^\top a}(t) + (\nu\ast s)(t) \\ 
&= \sum_i w_i(\epsilon(t-d)\ast s_i)(t) + (\nu\ast s) = \bm{w^\top a}(t) + (\nu\ast s)(t),
\end{align*}
where $\ast$ represents the convolution operation and $\epsilon(\cdot)$ is a spike response kernel that does not take into account delays.
Let $\vartheta$ be the spiking threshold: an output spike is generated when the membrane potential $u(t)$ reaches $\vartheta$, more formally: 
\begin{align*}
&f_s(u):u\rightarrow s,\ s(t):=s(t) + \delta(t-t^{(f+1)}) \\
&\text{where} \ \ t^{(f+1)} = \min\{t: u(t) = \vartheta, t > t^{(f)} \}.
\end{align*}

\subsection{Synaptic delays}

To achieve pure delay training, axonal delays are not sufficient since they lack expressivity. Training an axonal delay means in practice that, starting from a neuron in a layer $i$, the delay applied to all the neurons of the following layer $i+1$, is exactly the same, i.e. we have a fraction of the trainable parameters that we have when training the synaptic delays.
 Our implementation is prototypical and has not been optimized for a CUDA execution. Notice however that the number of \textit{trainable} parameters in a weight-based and \textit{synaptic delay-based} network is exactly the same; we expect therefore that a lower-level implementation would perform similarly to \citep{slayer} in terms of training time.
In the simulated environment, the inference is badly influenced by the double operation needed in a synapse (application of delay + multiplication), but we expect advantages in a hardware context. 

Taking into consideration the synaptic delays, the formulation is simply extended as: 
\begin{equation*}
u(t) = \sum_i w_i(\epsilon_{d_i}\ast s_i)(t) + (\nu\ast s)  = \sum_i w_i(\epsilon(t-d_i)\ast s_i)(t) + (\nu\ast s).
\end{equation*}
In all our experiments, the spike response kernels were:
\begin{align*}
    \epsilon (t) &= \frac{t}{\tau_s} \text{exp}(1-\frac{t}{\tau_s})\Theta (t), 
    \nu (t) &= \-2 \vartheta \text{exp}(1-\frac{t}{\tau_r}) \Theta(t),
\end{align*}
where $\Theta (t)$ is the Heaviside step function, although the formulation is independent of the chosen kernel.
Real-valued delays were stored for each synapse during training, but only the quantized values were used during inference. Quantization was obtained by a simple round \textit{down} to the nearest allowed number. With a simulation timestep of $1$ms, this meant that if we had a synaptic delay of, say, $4.421ms$, the spike is delayed by $4ms$.
A stochastic rounding \citep{stocastic} method was tried, but did not lead to improvements and led to an increase to compute time; it was considered therefore not worthy at this stage.

\subsection{Loss and spike target}
For all the experiments, a spike time-based loss was used. 
For a target spike train $\hat{s}(t)$, for a time interval $[0, T]$, the loss function was defined as:
\begin{equation*}
E=\int^T_0 L(s^{(n_l)},\hat{s}(t))\,dt = \frac{1}{2}\int_0^T (e^{n_l}(s^{(n_l)}(t), \hat{s}(t)))^2 dt,
\end{equation*}
where $ L(s^{(n_l)}(t),\hat{s}(t)) $ is the loss at time instance $t$ and $e^{(n_l)}(s^{(n_l)}, \hat{s}(t))$ is the error signal at the final layer. 

Finally, the error signal was: 
\begin{equation*}
e^{(n_l)}(s^{(n_l)}, \hat{s}(t)) = \epsilon *(s^{(n_l)}(t) - \hat{s}(t)) = a^{(n_l)}(t) - \hat{a}(t).
\end{equation*}
For the classification tasks, the target spike train was specified as the neuron corresponding to the correct class, spiking for the whole simulation period.
The class was inferred by looking at the first neuron that spikes; in the case where more than one neuron spikes, we infer the class of the neuron that, at parity of arrival time, has the most spikes.

\subsection{Data encoding}
For the image-based classification task we converted the non-temporal deep learning datasets to spiking encodings.
In this case, because the trained parameters (the delays) work intrinsically in a temporal domain, we opt for temporal encoding. 
Although a temporal coding similar to what is done in \citep{s4nn} was tried (i.e. for each pixel, a spike temporally placed in a proportional way to the pixel intensity), we found that a simpler encoding was more effective in terms of accuracy performance for our setup.
The strategy that we used in practice is one where, for each pixel, if the greyscale value is higher than 127 we get a spike, and otherwise we don't get a spike.

\subsection{Training delays}

\begin{figure}[h!]
  \includegraphics[scale=0.35]{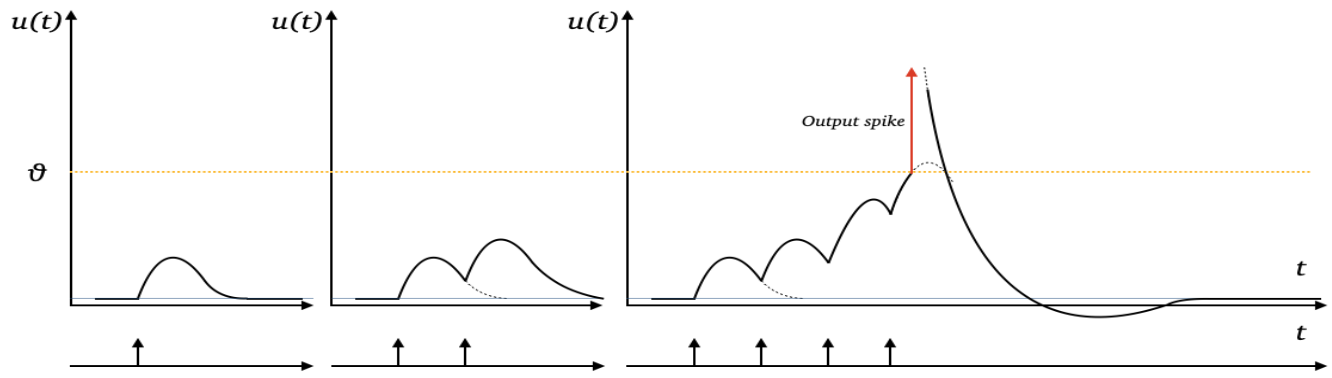}
  \caption{Spiking neuron dynamics. As a neuron receives a spike, a change in the membrane potential is obtained. As the membrane potential reaches a threshold $\vartheta$ a spike is generated}
  \label{fig:dynamics}
\end{figure}

\begin{figure}[h!]
  \includegraphics[scale=0.35]{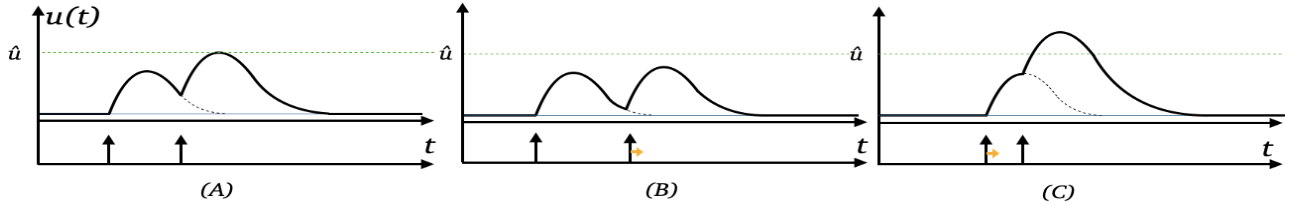}
  \caption{Delay dynamics. Demonstration of how changing the delay of a spike can change the amplitude of the membrane potential of a receiving neuron}
  \label{fig:delay_dynamics}
  \vspace{-4mm}
\end{figure}

Training delays is a fundamentally different operation, semantically, than training weights. 
Here we attempt to give an intuitive explanation at a high level, of what the neural dynamics can be. 
The explanation applies generally to all neuronal models.
The spikes cause an increase of the membrane potential $u(t)$ in the receiving neuron -- see Figure~\ref{fig:dynamics}. 
As the membrane potential reaches a threshold $\vartheta$, the receiving neuron will output a spike. 
Training synaptic weights act on how much a spike will impact the magnitude of the membrane potential: this is a direct action on the amplitude of the receiving spike \citep{neuralDynamics}.
Training delays, on the other hand, act on when (or if) a spike is generated, following different dynamics.
Remembering that generation of a spike is the consequence of a change in amplitude, we consider the latter, as the rest follows naturally. 
Considering Figure \ref{fig:delay_dynamics}: 
(A) shows a possible dynamics of a neuron receiving two spikes. 
Consider the magnitude $\hat{u}$ as the maximum value of the membrane potential $u(t)$ as the consequence of the spikes in this case, we will show how we can obtain a lower or higher maximum membrane potential value compared to $\hat{u}$. 
If we want to use delays to change the amplitude of the membrane potential, in the case depicted, we have two possibilities, delay the first spike, or the second. 
We can see that, delaying the second spike leads to the silhouette of the membrane potential never reaching $\hat{u}$ (Figure \ref{fig:delay_dynamics}.B). Delaying the first spike, leads to a $u(t)$ function that surpasses $\hat{u}$ (Figure \ref{fig:delay_dynamics}.A).

\subsection{Training procedure}

All our experiments were on a fully connected network with one hidden layer of 800 neurons (784-800-10). 
The optimization was done with the Adam \citep{adam}, with an initial learning rate of \num{0.01} and a batch size of \num{32}.
The training set was split to train and validation with an 80\%, 20\% split, and we report the test set accuracy in the tables.
Weights were initialized following a normal distributions: $\mathcal{N}(0.0571, 0.5458)$ for the first layer and $\mathcal{N}(-0.5244, 1.0490)$ for the second layer and scaled by a factor of $x10$ as in~\citep{slayer}, based on analysing the weight distributions of weight-trained networks

The weights are then scaled by a factor of $x10$, following the specifications of SLAYER. 
For the initialization of the \textit{constrained weights}, the same strategy was used, with the addition that, before applying the multiplicative factor, we apply the following, simple rule: 
let $w \in \mathbb{R}$ be a weight of the network. 
Let $\bar{w}$ be the quantized version of the weight. 
Then, $\bar{w} = round(w)$ where $round(\cdot)$ rounds $w$ to the nearest integer in the set $\{-1, 0, 1\}$. 

For the delays initialisation, we kept the same strategy used in~\citep{slayer}: A uniform random initialisation between 0 and 1.
In practice, this means that initially, the applied delay will be 0 in the forward pass, but non-zero initialization allows the avoidance of gradient problems in the backward pass. 
In some trials, a random delays initialisation in $[0,\, n]$ with different values for $n\in\mathbb{R}$ was also tried, but it did not lead to any performance improvements.

\subsubsection*{Parameters of the simulation}
All simulations have a duration of \qty{10}{ms}, and time is quantized with \qty{1}{ms} precision.
The spiking threshold was set to $\vartheta = \qty{10}{mV}$, the time constant to $\tau_s = \qty{1}{ms}$ and the refractory time constant to $\tau_r = \qty{1}{ms}$.


\section{Results}

We test pure delay training on two datasets, the MNIST~\citep{MNIST} hand-written digits dataset and the Fashion-MNIST~\citep{fashionMNIST} dataset.
We compare the results with weight training, and other similar methods in literature.

\subsection{MNIST}

\begin{figure}
  \includegraphics[scale=0.28]{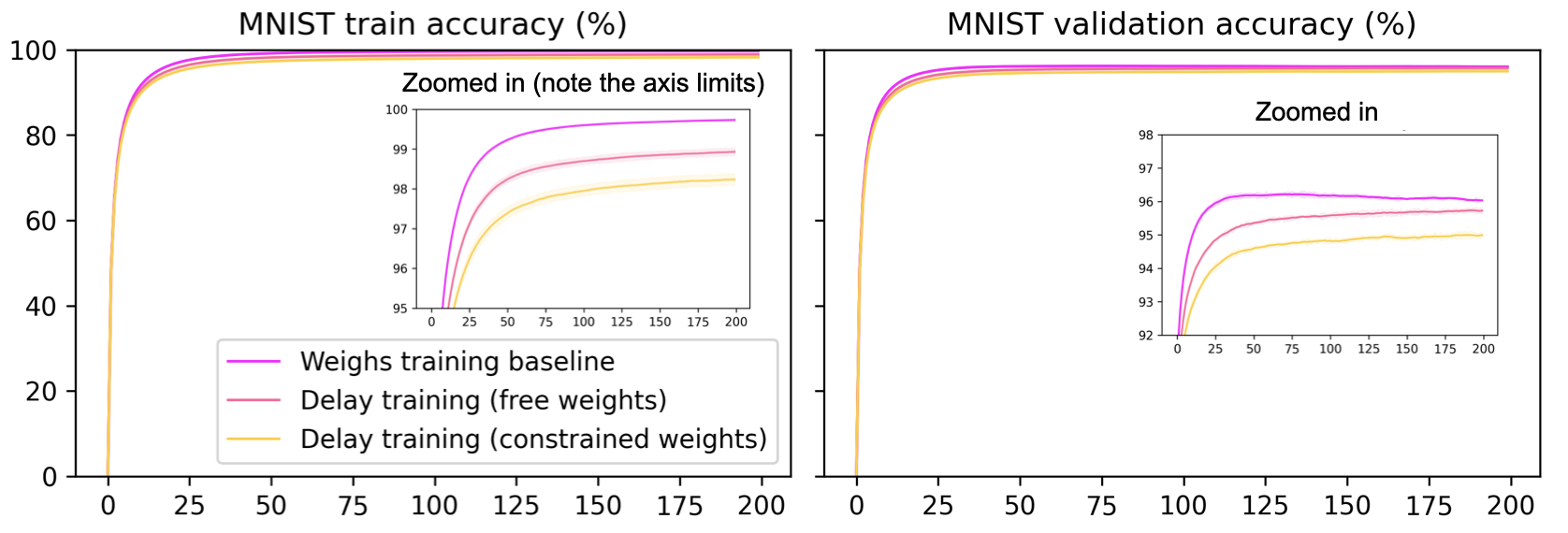}
  \caption{MNIST learning curve. \\Learning curves (accuracy) for training and validation sets. Full scale and zoomed version (inset).
  Without additional regularizers, delay training is more resilient to overfitting. 
  }
  \label{fig:MNIST_charts}
\end{figure}

We show in Figure~\ref{fig:MNIST_charts} the learning curves for our experiments. 
In the case of MNIST, we can see how the weights baseline fits the training dataset quickly, and our two methods do not achieve the same training accuracy. 
However, viewing the validation accuracy curve, we can see how, without additional regularizers, weights training overfits, in contrast to our method. 
With both free and constrained weights initialisation, we  achieve an accuracy improvement over \citep{hazan2022memory}, which is the only work that attempts a similar approach to ours (albeit not fully supervised).
Our goal here is to demonstrate a proof-of-concept that pure delay training is competitive rather than achieve state-of-the-art accuracy.

\begin{table}
\resizebox{\columnwidth}{!}{%
\begin{tabular}{lcccc}
\toprule
 & \textbf{Coding} & \textbf{Training Method} & \textbf{Neuron model} & \textbf{Accuracy (\%)} \\ 
 \midrule
S4NN \citep{s4nn} & Temporal & Backprop & IF & 97.4 \\ 
Memories with delays \cite{hazan2022memory} & Temporal & STDP & LIF & 93.5 \\ 
\midrule
ANN-fc (*) & - & Backprop & ReLU & $98.1 \pm 0.1$ \\ 
SLAYER weights baseline (*) & Temporal & Backprop & SRM & $96.1 \pm 0.1$ \\ 
\midrule
Delay training \\ w/ random free weights (ours) & Temporal & Backprop & SRM & $95.6 \pm 0.1$ \\ 
Delay training \\ w/ constrained weights (ours) & Temporal & Backprop & SRM & $94.9 \pm 0.1$ \\ 
\bottomrule
\end{tabular}
}
\caption{MNIST results. \\ Comparison with other related methods. 
All networks are fully connected networks (no CNNs). 
The accuracy values for our experiments correspond to the test evaluation on the model that performed best on the validation set, reported as mean and standard deviation values over 3 runs. 
(*) denotes runs done by us. For the (non-spiking) ANN baseline, we used a fully-connected architectures and the same number of neurons as our SNN experiments without additional regularizers. 
 Only \cite{hazan2022memory} train delays in the network directly.}
\label{tab:mnist}
\vspace{-8mm}
\end{table}
\subsection{Fashion-MNIST}
\begin{figure}
  \includegraphics[scale=0.28]{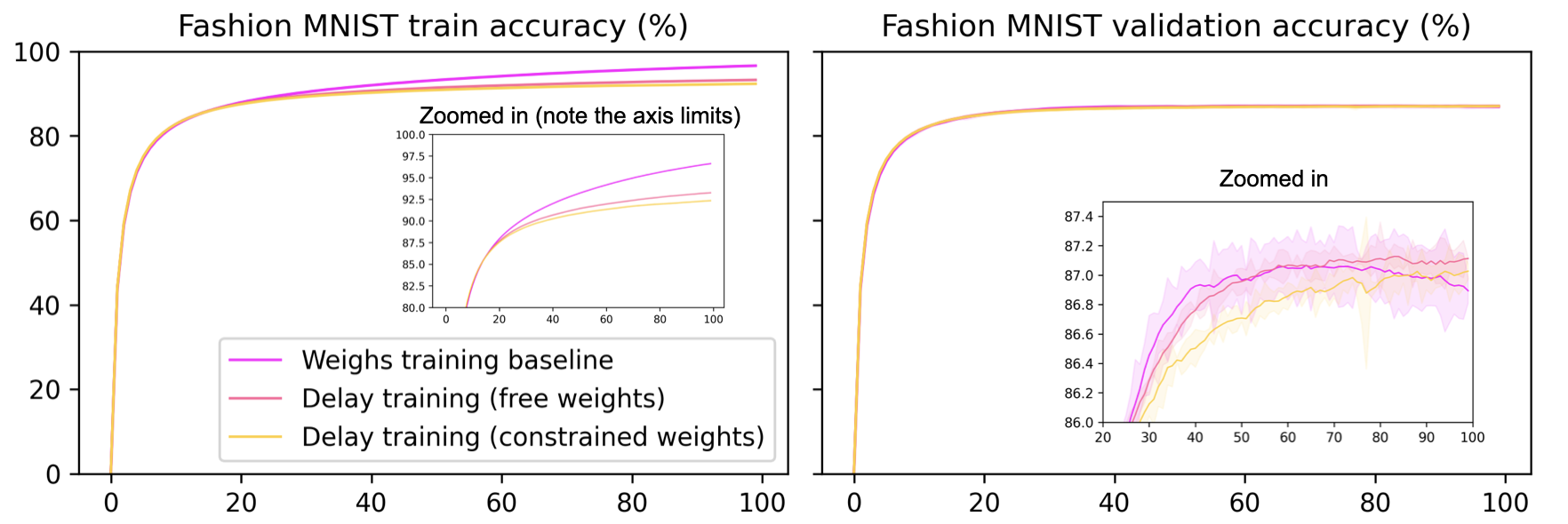}
  \caption{Fashion-MNIST learning curve.\\Weights training achieves high accuracy in the training set, but delay training performs better in the validation set.}
  \label{fig:Fashion-MNIST_charts}
  \vspace{-4mm}
\end{figure}

In Figure \ref{fig:Fashion-MNIST_charts}, we present the learning curves for the experiments on the Fashion-MNIST dataset. 
In this case, we observe how the overfitting on the training set for the weight training is even more evident than the previous case; weight training achieves over $97\%$ training accuracy on the training set, delay training stops at $92\%$.
We can see from the validation curves how, as epochs pass, our delay training method is more robust to overfitting, indeed, at the 100th epoch mark the weights training baseline has lower accuracy values than both our delay training methods. We also notice that our methods get very close to the weights training baseline for the best validation model, and are not far behind the non-spiking and basic ANN baseline.

\begin{table}
\resizebox{\columnwidth}{!}{%
\begin{tabular}{lcccc}
\toprule
 & \textbf{Coding} & \textbf{Training Method} & \textbf{Neuron model} & \textbf{Accuracy (\%)} \\ 
 \midrule
HM2BP \citep{capsnet} & Rate & Backprop & LIF & 89.0 \\
\midrule
ANN-fc (*) & - & Backprop & ReLU & $89.3\pm 0.2$ \\ 
SLAYER weights baseline (*) & Temporal & Backprop & SRM & $86.8 \pm 0.1$ \\ 
\midrule
Delay training \\ w/ random free weights (ours) & Temporal & Backprop & SRM & $86.6 \pm 0.02$ \\
Delay training \\ w/ constrained weights (ours) & Temporal & Backprop & SRM & $86.23 \pm 0.1$ \\ 
\bottomrule
\end{tabular}
}
\caption{Fashion-MNIST. 
(*) see Table \ref{tab:mnist}.
}
\label{tab:fashion_mnist}
\vspace{-8mm}
\end{table}

\section{Discussion}\label{sec:discussion}
We have demonstrated a proof of concept that training just the delays in a spiking neural network can work, as well as training weights.
We showed that on both MNIST and Fashion-MNIST datasets, pure delay training achieves task performance comparable to conventional weight training, with delay training having a slight advantage in not overfitting the dataset.
Moreover, we demonstrated that even when the weights are randomly initialised to ternary values ($\{+x,\, 0,\, -x\}$), the task performance of the networks remains good.

This is the first step toward understanding the computational power of delays in spiking neural networks for biology and machine learning.
Input and output encodings that use better temporal information and more precise delay training methods, and using pure delay-training for more powerful event-based models such as the EGRU~\citep{subramoney2023efficient} are potential ways to extend this in future work.

A forward pass in a spiking network using only delays with these ternary weights can also be implemented significantly more efficiently than a network that uses floating point weights.
In software, such a forward pass can use just matrix roll operations combined with addition/subtraction instead of multiply-accumulate.
In neuromorphic computing, our work suggests new ways of configuring the hardware to achieve extremely efficient inference. It might especially be relevant to analog and photonic~\citep{shastri2021photonics} neuromorphic devices, although significant savings can also be realised in other devices.
Overall, this work demonstrates a new paradigm of using time in and training spiking neural networks.

\begin{acks}

\begin{minipage}{0.07\textwidth}
\includegraphics[width=\linewidth]{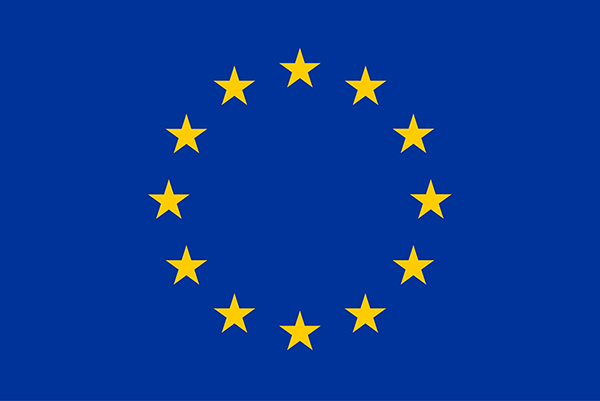}
\end{minipage}%
~\hfill%
\begin{minipage}{0.4\textwidth}
This work was co-funded by the European Union within the Erasmus+ traineeship program. 
\end{minipage}
\vspace{0.5em}

\noindent AS was funded by the Ministry of Culture and Science of the State of North Rhine-Westphalia, Germany during part of this work.
We would like to thank Andrew D. Bagdanov, Laurenz Wiskott and the Institute for Neural Computation at Ruhr University Bochum for institutional and infrastructural support.
\end{acks}

\vspace{-0mm}

\bibliographystyle{ACM-Reference-Format}
\bibliography{references}

\appendix









\end{document}